\pdfoutput=1
\documentclass[11pt]{article}

\usepackage{acl}

\usepackage{times}
\usepackage{latexsym}

\usepackage[T1]{fontenc}
\usepackage[utf8]{inputenc}

\usepackage{microtype}

\usepackage{hyperref}
\usepackage{url}

\usepackage{subcaption}
\usepackage{multirow}
\usepackage[shortlabels]{enumitem}
\usepackage{microtype}
\usepackage{graphicx}
\usepackage{booktabs} % for professional tables
\usepackage{hyperref}
\usepackage{ifthen}
\usepackage{cuted}
\usepackage{makecell}
\usepackage{tabularx}
\usepackage{multicol}
\usepackage{wrapfig}

\usepackage{amsmath}
\usepackage{amssymb}
\usepackage{amsthm}
\usepackage{dsfont}
\usepackage{multirow}
\usepackage[shortlabels]{enumitem}

\usepackage{hyperref}

\usepackage{amsmath,amsfonts,bm}

\def\eqref#1{equation~\ref{#1}}
\def\1{\bm{1}}

\def\vm{{\bm{m}}}

\def\vx{{\bm{x}}}

\DeclareMathAlphabet{\mathsfit}{\encodingdefault}{\sfdefault}{m}{sl}
\SetMathAlphabet{\mathsfit}{bold}{\encodingdefault}{\sfdefault}{bx}{n}

\usepackage[ruled,boxed,lined]{algorithm2e} % linesnumbered
\let\oldnl\nl% Store \nl in \oldnl
\newcommand{\nonl}{\renewcommand{\nl}{\let\nl\oldnl}}% Remove line number for one line

\newcommand{\comment}[1]{}

\usepackage{ifthen}
\providecommand{\dodraft}{true}
\newboolean{isdraft}
\setboolean{isdraft}{\dodraft} 
\ifthenelse{\boolean{isdraft}}{%
	\newcommand{\tianyi}[1]{{\color{blue}{{\bf Tianyi}}$\to$#1}}
	\newcommand{\richard}[1]{{\color{red}{{\bf Richard}}$\to$#1}}
	\newcommand{\lehou}[1]{{\color{green}{{\bf Le}}$\to$#1}}
	\newcommand{\todo}[1]{{\color{purple}{{\bf TODO: #1} }}}
	\newcommand{\fixme}[1]{{\color{brown}{{\bf FIXME: #1} }}}
	\newcommand{\afterfix}[1]{{\color{red}{{\bf AFTERFIX: #1} }}}
	
} {
	
	\newcommand{\tianyi}[1]{}
	\newcommand{\richard}[1]{}
	\newcommand{\lehou}[1]{}
	\newcommand{\todo}[1]{}
	\newcommand{\fixme}[1]{}
	\newcommand{\afterfix}[1]{}
	
}

\title{Token Dropping for Efficient BERT Pretraining}

\author{Le Hou$^{1*}$~~~~~Richard Yuanzhe Pang$^{12\S*}$~~~~~Tianyi Zhou$^{13\S}$~~~~~Yuexin Wu$^{1}$~~~~~Xinying Song$^{1}$\\
\bf{Xiaodan Song$^{1}$~~~~~Denny Zhou$^{1}$} \\
$^{1}$ Google~~~~~~$^{2}$ New York University~~~~~~$^{3}$ University of Maryland, College Park \\
{\tt lehou@google.com},\ {\tt yzpang@nyu.edu}}

\begin{document}
\maketitle

{
\let\thefootnote\relax\footnote{$^{*}$~Equal contribution.} 
\let\thefootnote\relax\footnote{$^{\S}$~Work done at Google Brain.} 

}

\begin{abstract}

Transformer-based models generally allocate the same amount of computation for each token in a given sequence. We develop a simple but effective ``token dropping'' method to accelerate the pretraining of transformer models, such as BERT, without degrading its performance on downstream tasks. 
%In particular, 
In short, we drop unimportant tokens starting from an intermediate layer in the model to make the model focus on important tokens; the dropped tokens are later picked up by the last layer of the model so that the model still produces full-length sequences. We leverage the already built-in masked language modeling (MLM) loss to identify unimportant tokens with practically no computational overhead. In our experiments, this simple approach reduces the pretraining cost of BERT by 25\% while achieving similar overall fine-tuning performance on standard downstream tasks.

\end{abstract}

\section{Introduction}
\label{sec:intro}

Nowadays, the success of neural networks in a variety of NLP tasks heavily relies on BERT-type language models containing millions to billions of parameters. However, the pretraining process of these models is computationally expensive,  generating significant emission \citep{strubell-etal-2019-energy,patterson2021carbon}. In practice, there is the need to perform large-scale language model pretraining for diverse applications \citep{lee2020biobert,chalkidis-etal-2020-legal,zou2021pretrained,rogers-etal-2020-primer} in different languages \citep{antoun-etal-2020-arabert,sun2021ernie}.  In this paper, we develop a technique that significantly reduces the pretraining cost of BERT models \citep{devlin-etal-2019-bert} without hurting their test performance on a diverse set of fine-tuning tasks.

Recent efforts of efficient training involve mixed-precision training \citep{shoeybi2019megatron}, distributed training \citep{you2020large}, better modeling on rare words and phrases \citep{wu2021taking}, designing more effective and data-efficient pretraining objectives \citep{lan2020albert,clark2020electra,raffel2020t5}, progressive stacking \citep{gong2019efficient}, and so on. While these approaches contribute to efficient training with reduced computational cost, most of them focus on the model architecture or the optimization process.  

In this paper, we focus on a simple but efficient BERT-pretraining strategy that has been under-explored, i.e.,  ``token dropping,'' which removes the redundant tokens in each sequence that are less informative to training. 
Since not all tokens contribute equally to the output or the training objective, and the computational complexity of transformer-based models grows at least linearly with respect to the sequence length, shortening the input sequences can accelerate the training effectively. 

\begin{figure}[t]
\centering
\includegraphics[clip, trim={135 100 137 100}, width=0.99\columnwidth]{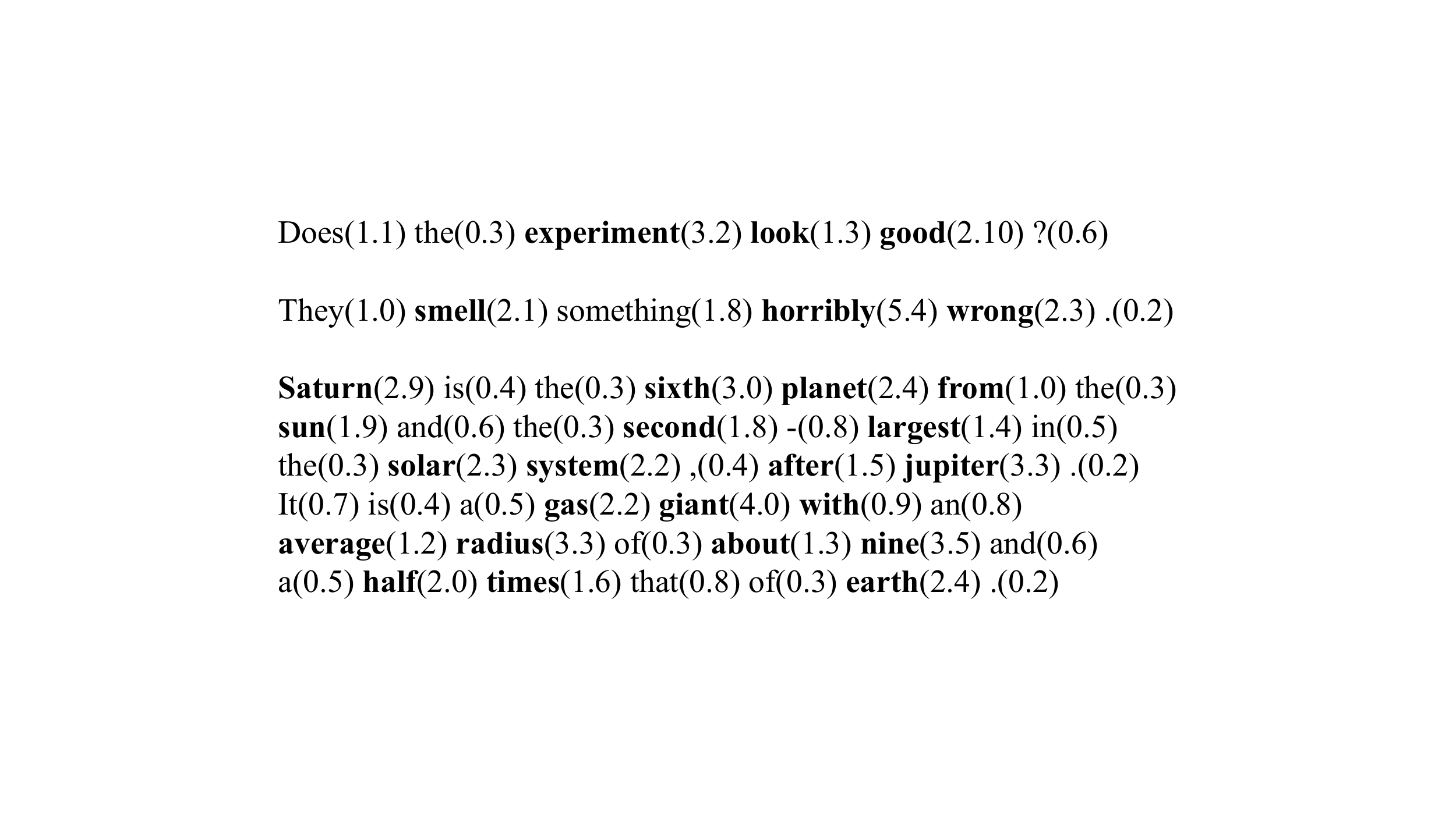}
\caption{Randomly selected example sentences with actual importance scores (cumulative losses, to be discussed in Section~\ref{sec:dropping}) from our model. % Numbers in parentheses are importance scores of the corresponding tokens. 
For efficient pretraining, tokens in bold are preserved in every BERT encoder layer, whereas other tokens are dropped for certain layers.}
\label{fig:drop_examples}
\end{figure}

Among existing studies, the depth-adaptive transformer approach aims to reduce the autoregressive inference time by allocating less computation on easy-to-predict tokens \citep{elbayad2020depth}. To improve the training efficiency, \citet{dai2020funnel} perform pooling on the embeddings of nearby tokens. 
However, directly dropping tokens during pretraining was not studied until very recently in faster depth adaptive transformer \citep{liu2021faster}, where the important tokens are identified either through (1) mutual information-based estimation between tokens and predefined labels or through (2) a separate BERT model that exhaustively computes the masked language model (MLM) loss for each token. On the contrary, we focus on accelerating the task-agnostic pretraining phase without requiring any labels or any computation by a separate language model. Specifically, we identify important tokens as the ones hard to predict by the model itself through its loss during training, which is adaptive 
to its training process and leads to practically no computational overhead. We show examples of dropped tokens in Figure \ref{fig:drop_examples}.

Recent approaches such as RoBERTa \citep{liu2019roberta} suggest packing input sequences. In this way, there are no \texttt{[PAD]} tokens, which makes it a non-trivial task to identify unimportant tokens. We identify unimportant tokens in each sequence with the smallest historical MLM loss (by taking the running average of the MLM loss of each token throughout the pretraining process). By removing them from intermediate layers of a BERT model during training, we save an enormous amount of computation and memory. We keep them in the first several layers as well as in the last layer so that they are still present in the model. Therefore, the inputs and outputs of BERT model are kept consistent with the conventional all-token training process. Without modifying the original BERT architecture or training setting, this simple token-dropping strategy trains intermediate layers mainly on a few important tokens. As demonstrated in our experiments, models pretrained in this way generalize well on diverse downstream tasks with full sequences.

To summarize, our contributions are as follows. (1) We show that BERT models can be pretrained with only a subset of the layers focusing on important tokens. Even though the model is trained on sub-sequences of important tokens only, it generalizes well to full sequences during fine-tuning on downstream tasks. (2) We identify important tokens through the pretraining process by exploring the training dynamics, with minimal computational overhead and without modifying the model architecture. (3) We show that our token dropping strategy can save 25\% of pretraining time while achieving similar performance on downstream tasks. Code is available at \url{https://github.com/tensorflow/models/tree/master/official/projects/token_dropping}.

\section{Prerequisites}
\label{sec:prereq}

\subsection{Sequence Packing}
\label{sec:packing}

Recall that a sequence in BERT consists of two sentences as well as the ``classification'' token \texttt{[CLS]} and the ``separator'' token \texttt{[SEP]}. If the resulting number of tokens is smaller than 512, then padding tokens are added to ensure that each sequence is exactly 512-token long.  

We decide to use sequence packing \citep{liu2019roberta} so that there would be no \texttt{[PAD]} symbols, throughout the paper. 
We also remove the next-sentence prediction training criteria as well. 
The rationale for using sequence-packing is two-fold. First, sequence packing provides a competitive baseline in terms of pretraining efficiency \citep{so2019evolved,liu2019roberta,kosec2021packing,zhang2021reducing}. Second, using sequence-packing can stress-test our algorithm under the absence of padding symbols to see if it brings further improvements beyond dropping padding tokens: without sequence packing, our algorithm can label \texttt{[PAD]} as the unimportant tokens, 
which trivially improves pretraining efficiency; with sequence packing, however, our algorithm has to identify and drop real tokens as unimportant tokens to improve the efficiency.

\subsection{Multi-Head Attention}

Define $T$ to be the input sequence length and $d_k, d_v$ to be the size of each individual key vector and value vector, respectively. The multi-head attention function with $h$ attention heads is defined as:
\begin{align}
    & \mathrm{MultiHeadAttention}(Q, K, V) = \nonumber \\ 
    & \qquad \mathrm{concat}(H_1,\dots, H_h) W^O, \nonumber
\end{align}
where
\begin{align}
    H_i &= \mathrm{Attention}(QW_i^Q, KW_i^K, VW_i^V) \nonumber \\
    &= \mathrm{softmax}\left(\frac{(QW_i^Q)(KW_i^K)^\top}{\sqrt{d_k}}\right) VW_i^V. \nonumber
\end{align}
Use $d_{\text{model}}$ to denote the hidden size of the model (usually equal to $hd_k$). We have the following: 
\begin{align}
    & Q,K,V \in \mathbb{R}^{T\times d_{\text{model}}}, \nonumber \\
    & W_i^Q \in \mathbb{R}^{d_{\text{model}}\times d_k}, 
    \quad W_i^K \in \mathbb{R}^{d_{\text{model}}\times d_k}, \nonumber \\
    & W_i^V \in \mathbb{R}^{d_{\text{model}}\times d_v}, 
    \quad W^O \in \mathbb{R}^{ hd_v \times d_{\text{model}}}. \nonumber
\end{align}

\subsection{Feed-Forward Networks}

Besides the attention sub-layer, each BERT encoder layer also contains the feed-forward sub-layer (or ``feed-forward network,'' abbreviated as FFN). Each FFN is a position-wise function: it is applied to each position of the input, identically. 

The input to FFN is a matrix $\in \mathbb{R}^{T \times d_{\text{model}}}$. The input will be transformed by a two-layer perceptron (with ReLU activation in between) into an output matrix $\in \mathbb{R}^{T \times d_o}$. In \citet{vaswani2017attention}, $d_o = d_{\text{model}}$ and the hidden size of the intermediate layer $d_{ff} = 4\cdot d_{\text{model}}$ due to empirical investigation.

\section{Token-Dropping}
\label{sec:methods}

\begin{figure}[t]
\centering
\includegraphics[clip, trim={25 50 25 50}, width=0.97\columnwidth]{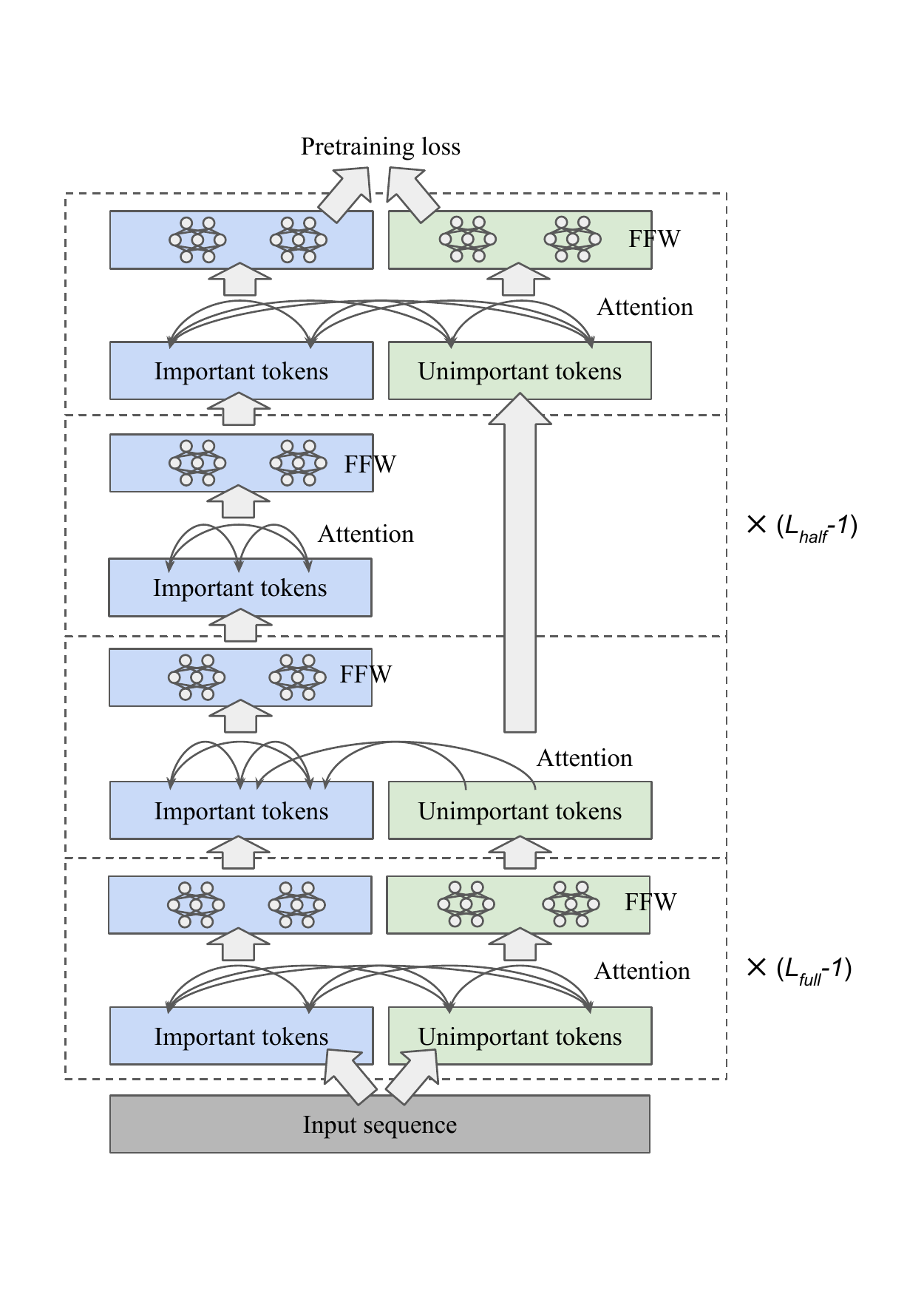}
\caption{Illustration of the token dropping method. The first several layers and the last layer process all tokens (the order of tokens are preserved in our implementation; the separation of important and unimportant tokens in the figure is for illustration purposes only). The middle layers only process important tokens. The important tokens are identified based on the historical MLM loss of each token: we maintain the running average of the MLM loss of each token.}
\label{fig:method_illustration}
\end{figure}

Suppose the input sequence contains 512 tokens. Having 512 hidden states (corresponding to 512 tokens) after each encoder layer may not be necessary, given that certain words may never heavily influence the meaning of the sentence.\footnote{Relatedly, \citet{zhang2019all} have shown in computer vision, using fully-connected networks and convolutional neural networks, that certain layers called ``ambient'' layers can be reset with almost no negative consequence on performance, while other layers called ``critical'' layers are necessary.} 
Moreover, removing unimportant tokens in intermediate layers would produce a ``dropout'' effect, so that our network would learn to reconstruct the masked words with more noisy hidden states. 
Therefore, we decide to allocate the full amount of computation only to important tokens. 
Figure \ref{fig:method_illustration} gives an illustration of where the unimportant tokens are dropped in a BERT model.

\subsection{Stage-1 Pretraining} 

Each row of query ${Q}$, key ${K}$, and value ${V}$ in a self-attention module in each transformer encoder layer corresponds to a single token. Suppose $L_f=L_{\text{full}}$ is the set of layers whose input covers all the tokens;\footnote{In this paper, ``input covers all the tokens'' means that the query, key, and value metrics in the layer have $T$ rows, so no rows are discarded from the matrices.} $L_h=L_{\text{half}}$ is the set of layers whose input only cover a proper subset of tokens. 

\paragraph{Separation.} During stage-1 pretraining, if the layer $l\in L_f$ and the next layer $l+1\in L_h$, then we remove the rows in ${Q}$ corresponding to the unimportant tokens for layer $l+1$ but keep ${K}$ and ${V}$ intact. After the removal, we have $Q \in \mathbb{R}^{M \times d_{\text{model}}}$ where $M$ is the number of important tokens. We also have $K, V\in \mathbb{R}^{T \times d_{\text{model}}}$ where $T$ is the input sequence length.\footnote{In practice, using TensorFlow, the separation step in stage-1 pretraining and the merge step can be done using the function \texttt{tf.gather()}. The number of important tokens for different sequences has to be the same in order to use modern accelerators like TPUs. Using sparse tensors can address the issue of having a different number of important tokens, but sparse tensor related operations in practice are slow.}

Suppose $l'$ is the first layer above layer $l+1$ such that $l' \in L_f$. Suppose $l+2 \in L_h$. Then, for layers $l+2$, \dots, $l'-1$, we have $Q, K, V \in \mathbb{R}^{M \times d_{\text{model}}}$, which means that their rows correspond to only the important tokens.

\paragraph{Merging.} Given that $l'$ is the first layer above layer $l+1$ such that $l' \in L_f$, before layer $l'$, we merge the hidden states corresponding to the unimportant tokens (taken from the outputs of layer $l$) with the hidden states corresponding to the important tokens (taken from the outputs of layer $l'-1$). We keep the order of hidden states consistent with the order of the input tokens.

\paragraph{Alternatively: token passing instead of token dropping.} In layers where unimportant tokens are dropped, the input to the layers effectively corresponds to partial and incoherent sentences. We thus attempt the token passing approach, which can ensure that the input to such layers corresponds to complete and coherent sentences. Token passing is described as follows.

In layers $l+1, \dots, l'-1 \in L_h$, we can keep the rows of ${K}$ and ${V}$ corresponding to the unimportant tokens. More specifically, the rows of $K$ that correspond to important tokens come from the hidden states outputted by the previous encoder layer. The rows of $K$ that correspond to unimportant tokens come from the hidden states outputted by layer $l$. This procedure results in $Q \in \mathbb{R}^{M\times d_{\text{model}}}$ and $K, V \in \mathbb{R}^{T \times d_{\text{model}}}$ for layers $l+1, \dots, l'-1$. See Section~\ref{sec:results} for empirical studies.

\paragraph{Determining $l$ and $l'$.} We leave details on determining $l$ and $l'$ to later sections. Empirically, $l = \frac{L_E}{2} - 1$ and $l' = L_E - 1$ consistently lead to good performance, where $L_E$ is the total number of encoder layers. For instance, if $L_E=12$, then the full layers in $L_f$ (i.e., layers in which the query, key, and value matrices all have $T$ rows) would be layers 1 through 5 as well as layer 12.

\begin{figure}[t]
\centering
\includegraphics[clip, trim={50 20 30 20}, width=0.9\columnwidth]{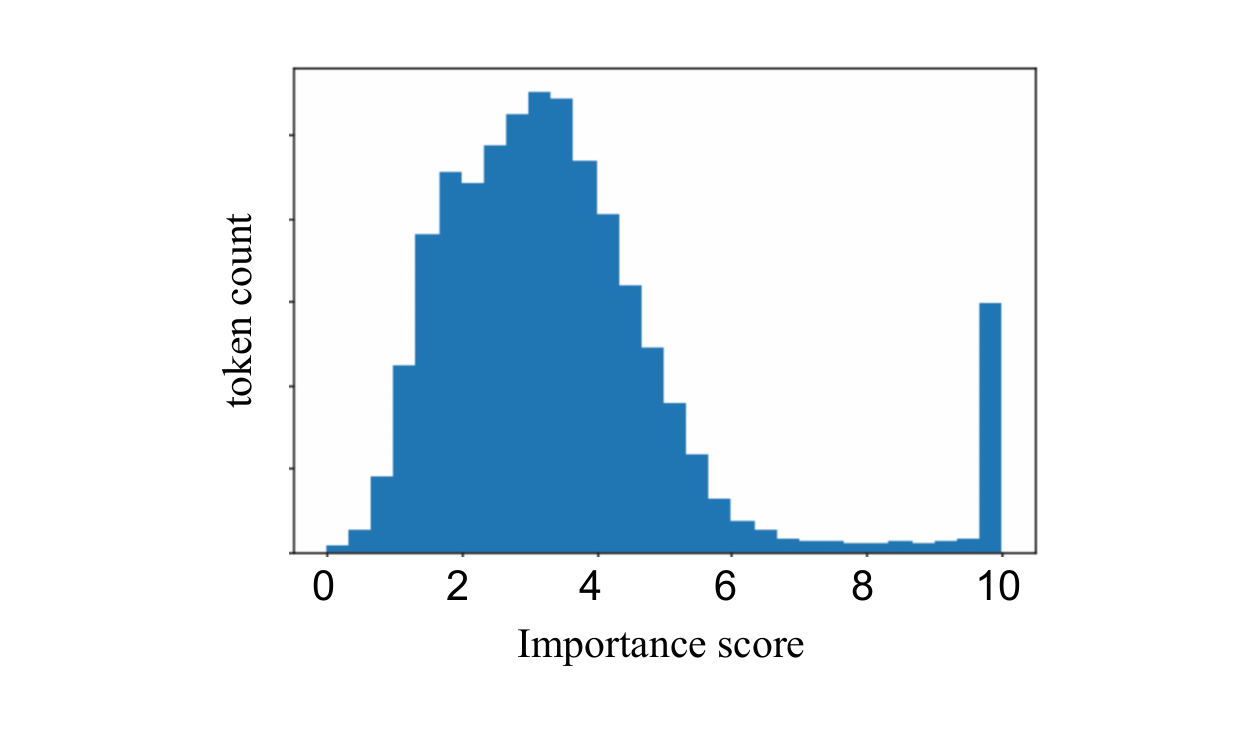}
\caption{The distribution of importance scores (cumulative losses) derived from the pretraining process according to Equation \ref{eq:importance_score}. If a token has not been masked before, it has the default cumulative loss of 10.}  
\label{fig:importance_score_distribution}
\end{figure}

\subsection{(Optional) Stage-2 Pretraining} 

At test-time or when we fine-tune on downstream tasks, all the encoder layers are full layers, meaning we do not do any token dropping. Given the mismatch between the neural network in stage 1 and the neural network used for fine-tuning and test-time, during stage 2, we simply pretrain using the full model (i.e., all tokens passing through all layers). Stage-2 pretraining requires only a smaller number of steps, compared to stage-1 pretraining. 
However, stage-2 pretraining turns out to be unnecessary, which we discuss in later sections. 

\subsection{Identifying Important Tokens}
\label{sec:dropping}

In this subsection, we elaborate on which tokens to drop (i.e., which corresponding rows to discard in the query, key, and value matrices) in a given sequence. First, we never drop special tokens including \texttt{[MASK]}, \texttt{[CLS]}, and \texttt{[SEP]}. In other words, we always treat these tokens as important tokens. Recall that we use sequence packing in all of our experiments, unless noted otherwise. Therefore, there are no padding tokens \texttt{[PAD]}.\footnote{If we do not use sequence packing, we would always drop the \texttt{[PAD]} tokens.}

We introduce two approaches for identifying important tokens in the following sub-sections. In the ablation studies (Section~\ref{sec:different-models}), we will introduce more straightforward approaches as baselines.

\subsubsection{Dynamic Approach: Cumulative-Loss-Based Dropping}

\paragraph{Updating the cumulative loss vector.} We use a vector $\vm \in \mathcal{R}^{|\mathcal{V}|}$ to approximate the ``difficulty'' of learning a specific token in the vocabulary $\mathcal{V}$. The vector $\vm$ is updated throughout the pretraining. Recall that BERT pretraining involves the masked language modeling (MLM) objective, where the model is asked to predict the tokens of the masked-out input tokens. Suppose $n$ tokens in a sequence are masked out, then we would obtain $n$ MLM negative log-likelihood (NLL) losses. 
For each token, we update the corresponding entry in the cumulative loss vector as follows: 
\begin{align}
\label{eq:importance_score}
    m_i \leftarrow \beta \cdot m_i + (1-\beta) \cdot \ell_i, 
\end{align}
where $\ell_i$ is the NLL loss that corresponds to the token $i$ and $\beta \in (0,1)$ is a coefficient that is close to 1. In particular, we never update the cumulative losses corresponding to the aforementioned special tokens (\texttt{[MASK]}, \texttt{[CLS]}, and \texttt{[SEP]}). The losses for those tokens are set to a large number such as $10^4$. If there are padding tokens in the sequence, then we set the loss to a negative number $-10^4$ so that we can ensure that the padding token has the smallest loss---given that NLL loss is always non-negative for all other tokens.

\paragraph{Deciding which tokens are unimportant.} We drop the rows in the query, key, and value matrices corresponding to the unimportant tokens. To decide which tokens will be treated as unimportant ones, given a sequence of 512 tokens, we simply look up the 512 corresponding cumulative losses using $\vm$, and label the tokens that correspond to the smallest cumulative losses as unimportant tokens. In other words, suppose we have a sequence $\vx = (x_1, x_2, \dots, x_T)$ where $T$ is the sequence length. Use $[T]$ to denote $\{1, 2, \dots, T\}$. Suppose $\sigma: [T] \to [T]$ is a function such that $x_{\sigma(1)}, x_{\sigma(2)}, \dots, x_{\sigma(T)}$ are the tokens 
sorted in decreasing order of the aforementioned cumulative loss. Then, we are treating $x_{\sigma(1)}, \dots, x_{\sigma(M)}$ as important tokens (i.e., the tokens to keep), where $M$ is a positive integer (e.g., $M=\mathrm{int}(T/2)$), and we are treating $x_{\sigma(M+1)}, \dots, x_{\sigma(T)}$ as unimportant tokens.\looseness-1 

\paragraph{Optionally: adding randomness.} We can optionally assign every token with a nonzero probability to be selected as an important token, which can potentially make the model generalize well on full sequences. For example, let $J = \mathrm{int}(0.05T)$, given tokens $x_{\sigma(1)}, x_{\sigma(2)}, \dots, x_{\sigma(T)}$ as described in the previous paragraph, we replace the last $J$ important tokens  $x_{\sigma(M-J+1)}, \dots, x_{\sigma(M)}$ with $J$ tokens randomly chosen from $x_{\sigma(M-J+1)}, \dots, x_{\sigma(T)}$. Then, the $J$ randomly chosen tokens will be treated as important tokens. In later sections, we will empirically investigate whether the randomness is helpful.

\subsubsection{Static Approach: Frequency-Based Dropping}
\label{sec:method-freq}

Before the start of pretraining, a simple program counts the number of occurrences of each token in the vocabulary $\mathcal{V}$. During pretraining, given a sequence, suppose there are $s$ special tokens. This approach assigns the special tokens as well as the $M-s$ tokens that correspond to the lowest frequency as important tokens, where $M$ is the target number of important tokens in a sequence. It treats the rest of the tokens as unimportant tokens.

\section{Experimental Details}
\label{sec:exp}

\subsection{Datasets}

\paragraph{Pretraining.} For pretraining, we use the same dataset as BERT: the BooksCorpus dataset \citep{zhu2015aligning} and the English Wikipedia dataset. We use the sequence-packed version of the dataset (Section~\ref{sec:packing}) so as to ensure that we have to drop meaningful tokens instead of the \texttt{[PAD]} tokens. 

\paragraph{Downstream tasks.} The BERT models are fine-tuned on GLUE tasks \citep{wang-etal-2018-glue} whose datasets are on the larger end. We only use the 6 largest GLUE datasets: MNLI, where we use MNLI-m to denote MNLI-matched and MNLI-mm to denote MNLI-mismatched \citep{williams-etal-2018-broad}, QNLI \citep{rajpurkar-etal-2016-squad}, QQP\footnote{https://quoradata.quora.com/First-Quora-Dataset-Release-Question-Pairs \label{qqp-source}}, SST \citep{socher-etal-2013-recursive}, and the GLUE diagnostics set AX \citep{wang-etal-2018-glue}. Additionally, we also experiment on the question answering datasets: SQuAD v1.1 \citep{rajpurkar-etal-2016-squad} and SQuAD v2.0 \citep{rajpurkar-etal-2018-know}. The evaluation metric for each task can be found in Table~\ref{tab:result-base}.

\subsection{Methods Tested}
\label{sec:different-models}

By default, the total training steps for each model is 1 million, using the settings in Section~\ref{sec:hyperparams}. We experiment with the following models. First, we have the baseline models.
\begin{itemize}
    \item \textit{baseline (no sequence packing)}: The original BERT with the non-sequence-packed input. 
    \item \textit{baseline}: The original BERT with the sequence-packed input.
    \item \textit{baseline (75\% steps)}: The original BERT with the sequence-packed input but only trained for 75 \% of the steps. This baseline is trained using a similar amount of computation as our proposed token dropping methods.
\end{itemize}

Next, we have the following methods that aim to save pretraining time. For token dropping methods, we drop 50\% of the tokens (unless mentioned otherwise) in order to compare with the average pooling method \cite{dai2020funnel} which reduces the sequence length by half.
\begin{itemize}
    \item \textit{token drop}: We perform stage-1 pretraining using the cumulative-loss token-dropping for 1M steps.
    \item \textit{token drop (rand)}: Similar to the ``token drop'' method, except that we randomly drop 50\% non-special tokens in a sequence, instead of dropping unimportant tokens. Special tokens like \texttt{[CLS]} and \texttt{[SEP]} are not dropped.
    \item \textit{token drop (half-rand)}: It is similar to the ``token drop'' method except adding extra randomness to the important token selection, as introduced in Section~\ref{sec:dropping}. This half-random method can be viewed as a combination of ``token drop'' and ``token drop (rand).''
    \item \textit{token drop (layer rearranged)}: It is similar to the ``token drop'' method except moving the last layer that processes all tokens to the beginning of the model. In other words, the layers in Figure \ref{fig:method_illustration} are rearranged such that full-sequence layers are only at the bottom.
    \item \textit{token drop (freq)}: Similar to the ``token drop'' method, except that we identify important tokens using the frequency-based token-dropping scheme, as discussed in Section~\ref{sec:method-freq}.
    \item \textit{token avg}: Similar to the ``token drop'' method, except that we use average-pooling to compress the sequences instead of ``token drop.'' Suppose layer $l \in L_f$ and the immediate next layer $l' \in L_h$, as described in Section~\ref{sec:methods}. Instead of dropping rows of the query, key, and value matrices, we apply average pooling with a window size of 2 and a stride of 2. In other words, suppose $q_1, \dots, q_T$ are the rows of the query vector. Then, the $T/2$ new query vectors are $(q_1 + q_2)/2, (q_3 + q_4)/2, \dots, (q_{T-1}, q_T)/2$, assuming that $T$ is an even number. This idea is introduced in Funnel transformer \cite{dai2020funnel}.
    \item \textit{token pass}: As discussed in Section~\ref{sec:methods}, we drop certain rows in the query, but we do not drop any row in the key and value matrices.
\end{itemize}

We also experiment with adding the optional stage-2 pretraining phase to the methods described above. In such cases, we first perform stage-1 pretraining for 900k steps and then stage-2 pretraining for 100k steps. To distinguish between the stage-1 only methods, we add {\textit{+ stage-2}} at the end of the method description.

\begin{table*}[t!]
\setlength{\tabcolsep}{2.2pt}
\centering
% \fontsize{11}{13.2}\selectfont
\small
% \resizebox{\textwidth}{!}{%
\begin{tabular}{>{\hangindent=0.9em\hangafter=1}m{4cm}cccccccc|ccc}
\toprule
 \multirow{3}{*}{Methods} & & \multicolumn{10}{c}{BERT-base}  \\
 \cline{3-12}
 \noalign{\smallskip}
 & &   AX   &  MNLI-mm & MNLI-m  & QNLI  & QQP & SST   & GLUE-avg & SQuAD    & SQuAD    & SQuAD \\
 & & (corr.)  &  (acc.)    &  (acc.)   & (acc.)  & (F1) & (acc.)  &          & -v1 (F1) & -v2 (F1) & -avg   \\
 \cline{1-12}
 \noalign{\smallskip}

baseline (no sequence packing) & & 76.36 & 84.61 & 84.28 & \textbf{91.56} & 90.94 & 95.73 & 87.25 & 90.11 & 78.89 & 84.50 \\
baseline & & 76.52 & 84.47 & 84.44 & 90.58 & 90.97 & 96.18 & 87.19 & 89.71 & 79.00 & 84.35 \\
baseline (75\%) & & 76.38 & 84.43 & 84.36 & 90.21 & 90.82 & 96.00 & 87.04 & 89.33 & 78.14 & 83.73 \\

\specialrule{.2pt}{1pt}{1pt}

\textbf{proposed} token drop & & \textbf{77.77} & \textbf{85.28} & \textbf{85.20} & 91.25 & \textbf{91.00} & 95.54 & \textbf{87.67} & 90.44 & \textbf{81.09} & \textbf{85.77} \\
token drop + stage-2 & & 77.70 & 84.91 & 85.04 & 91.40 & \textbf{91.00} & 95.98 & \textbf{87.67} & 90.32 & 79.90 & 85.11 \\
token drop (half-rand) & & 77.08 & 84.92 & 84.81 & 91.36 & 90.94 & 96.80 & 87.65 & 90.34 & 80.38 & 85.36 \\
token drop (half-rand) + stage-2 & & 77.25 & 85.19 & 84.89 & 91.52 & 90.67 & 94.94 & 87.41 & \textbf{90.47} & 79.81 & 85.14 \\
token drop (rand) + stage-2 & & 76.88 & 84.56 & 84.56 & 91.27 & 90.78 & 95.65 & 87.28 & 89.65 & 78.61 & 84.13 \\
token drop (freq) + stage-2 & & 76.19 & 84.35 & 84.27 & 91.05 & 90.80 & 96.48 & 87.19 & 89.38 & 77.32 & 83.35 \\
token avg + stage-2 & & 76.92 & 84.83 & 84.69 & 90.94 & 90.89 & \textbf{97.03} & 87.55 & 90.23 & 79.35 & 84.79 \\
token pass + stage-2 & & 77.04 & 84.58 & 84.86 & 91.36 & 90.89 & 95.67 & 87.40 & 89.98 & 79.85 & 84.92 \\
token drop (layer rearranged) + stage-2 & & 76.61 & 84.52 & 84.37 & 90.78 & 90.76 & 96.65 & 87.28 & 90.05 & 78.38 & 84.21 \\

\bottomrule
\end{tabular}
\caption{Evaluating different pretraining methods by finetuning pretrained models on downstream tasks. We pretrain BERT-base models on packed sequences of 512 tokens. Each number corresponds to the average of two different pretraining and finetuning runs (using different random seeds).
\label{tab:result-base}}
\end{table*}

\subsection{Model Architectures}

The BERT architectures are the same as the ones in \citet{devlin-etal-2019-bert}. We experiment on both BERT-base and BERT-large. For each BERT architecture, we train with two different input sequence lengths: 512 and 128. We use the sequence-packed input data, unless otherwise noted.

\subsection{Hyperparameters and Other Details}
\label{sec:hyperparams}

We use TPUv3 to pretrain the BERT models. The batch size of each pretraining step is 512. We train each BERT model for 1 million steps. We use the AdamW optimizer \citep{loshchilov2018decoupled}. We adopt a peak learning rate of $1e-4$ and use the linear decay scheduler for the learning rate.

We conduct extensive hyperparameter tuning for downstream tasks. For all GLUE tasks, we test different numbers of training epochs $\xi \in \{2,3,4,5,6,8,10\}$ and peak learning rate values $\eta \in \{5e-6,1e-5,2e-5,3e-5,4e-5\}$ using the baseline pretrained BERT model. $\xi \in \{3,6\}$ and $\eta \in \{1e-5,2e-5\}$ give the best overall results. Thus, for every pretrained model, we fine-tune on each individual GLUE task using the combinations of the two best $\xi$ and $\eta$ values (four settings in total) and take the best validation result.  
For SQuAD tasks, we test $\xi \in \{1, 2, 3, 4, 5, 6, 8\}$ and $\eta \in \{5e-5,6e-5,8e-5,1e-4,1.2e-4\}$ using the baseline pretrained BERT model and find out that $\xi \in \{4,8\}$ and $\eta \in \{2e-5,4e-5\}$ produce the best results overall. Thus, we fine-tune every model with these settings and report the best validation result.

We apply the linear decay learning rate schedule that ends with zero for all experiments. 
For each method, we pretrain two models with different random seeds. Then, these two models are fine-tuned separately on individual downstream tasks. We then report the averaged result as the final result for each task.

\section{Results}
\label{sec:results}

Table~\ref{tab:result-base} shows the ablation study. As mentioned, each number in the table corresponds to the average performance of two pretrained models (using different random seeds) that are then separately fine-tuned.

\subsection{Observations}

\paragraph{On whether stage-2 pretraining is useful.}

There is a mismatch between the neural network in stage-1 pretraining and the neural network used for fine-tuning and test-time. Therefore, we propose stage-2 pretraining where there is no token dropping so as to address the train-test mismatch. 
Comparing ``token drop'' with ``token drop + stage-2'' in Table~\ref{tab:result-base}, we see that the performance of the model trained without stage-2 pretraining and the model trained with stage-2 pretraining perform similarly. We hypothesize that the train-test mismatch can be easily addressed during downstream task fine-tuning. 

\paragraph{On determining which tokens are important.} 

Figure~\ref{fig:drop_examples} shows which tokens are labeled as important using three examples from our ``token drop'' model. 
Additionally, in Section~\ref{sec:dropping}, we propose to optionally replace the important tokens that have the lowest cumulative losses with unimportant tokens. Comparing ``token drop`` with ``token drop (half-rand)'' and ``token drop (rand)'' in Table~\ref{tab:result-base}, we see that adding randomness does not help. 
Finally, we see that the cumulative-loss-based dropping performs better than frequency-based dropping and random dropping. 
 
\paragraph{On how many tokens to drop. }

We report results with different token dropping percentages on training the BERT-base model in Table ~\ref{tab:how_many_to_drop}. We see that dropping more than 62.5\% of the tokens yield worse results. By default, our experiments drop 50\% of the tokens.

\paragraph{On determining which layers to drop.}

Comparing ``token drop (half-rand) + stage-2'' with ``token drop (layer rearranged) + stage-2,'' we can see that putting one full-sequence layer at the end of the model yields better results.

\paragraph{On token dropping vs. token passing.}

Comparing ``token drop + stage-2'' with ``token pass + stage-2,'' we see that passing the unimportant tokens instead of dropping them does not affect the performance. Recall that for layers where unimportant tokens are dropped, token dropping would make the input to such layers correspond to incoherent sentences, which could impact BERT's learning ability. However, we find that doing token passing makes pretraining slightly less efficient while providing no improvement on downstream performance. 

\paragraph{On token dropping vs. token averaging.}

Comparing ``token drop + stage-2'' with ``token avg + stage-2,'' we see that average pooling instead of dropping unimportant tokens yields slightly worse results. This means that our importance-driven token selection is more efficient than directly averaging embedding across every nearby token pair.

\begin{table*}[t!]
\setlength{\tabcolsep}{2.8pt}
\centering
% \fontsize{11}{13.2}\selectfont
\small
% \resizebox{\textwidth}{!}{%
\begin{tabular}{>{\hangindent=0.9em\hangafter=1}m{3cm}cccccccc|ccc}
\toprule
 \noalign{\smallskip}
  \multirow{2}{*}{Methods} & &   AX   &  MNLI-mm & MNLI-m  & QNLI  & QQP & SST   & GLUE-avg & SQuAD    & SQuAD    & SQuAD-avg \\
 & & (corr.)  &  (acc.)   &  (acc.)  & (acc.)  & (F1) & (acc.)  &          & v1 (F1) & v2 (F1) &   \\
 \noalign{\smallskip}
 \cline{1-12}
 \noalign{\smallskip}
\multicolumn{12}{c}{BERT-large, sequence length 128} \\
\specialrule{.2pt}{1pt}{1pt}
baseline                         & & \textbf{78.69} & \textbf{85.64} & \textbf{85.82} & 90.86 & \textbf{91.05} & 96.42 & 88.08 & 81.69 & 75.31 & 78.50 \\
\textbf{proposed} token drop     & & 78.61 & 85.42 & 85.46 & \textbf{91.39} & 90.64 & \textbf{97.98} & \textbf{88.25} & 82.91 & 75.18 & 79.05 \\
token drop + stage-2             & & 78.59 & 85.41 & 85.55 & 91.08 & 90.59 & 97.03 & 88.04 & 82.19 & \textbf{75.48} & 78.84 \\
token drop (half-rand)           & & 77.90 & 85.20 & 85.34 & 90.17 & 90.60 & 96.95 & 87.70 & \textbf{83.20} & 75.34 & \textbf{79.28} \\
token drop (half-rand) + stage-2 & & 78.51 & 85.49 & 85.56 & 91.33 & 90.67 & 97.21 & 88.13 & 82.91 & 74.76 & 78.84 \\
\specialrule{.2pt}{1pt}{1pt}
\multicolumn{12}{c}{BERT-large, sequence length 512} \\
\specialrule{.2pt}{1pt}{1pt}
baseline                         & & \textbf{81.94} & \textbf{87.56} & \textbf{87.97} & \textbf{93.78} & 91.36 & 96.89 & \textbf{89.92} & \textbf{93.30} & 85.89 & \textbf{89.59} \\
\textbf{proposed} token drop     & & 81.48 & 87.00 & 87.23 & 92.91 & 91.24 & \textbf{97.75} & 89.60 & 92.80 & \textbf{85.92} & 89.36 \\
token drop + stage-2             & & 81.18 & 87.34 & 87.53 & 93.46 & \textbf{91.46} & \textbf{97.75} & 89.79 & 92.88 & 85.69 & 89.28 \\
token drop (half-rand)           & & 80.73 & 87.19 & 87.22 & 93.52 & 91.21 & 97.69 & 89.59 & 92.67 & 85.19 & 88.93 \\
token drop (half-rand) + stage-2 & & 80.86 & 87.03 & 87.56 & 92.75 & 91.05 & 97.48 & 89.45 & 92.48 & 85.11 & 88.80 \\
\specialrule{.2pt}{1pt}{1pt}
\multicolumn{12}{c}{BERT-base, sequence length 128} \\
\specialrule{.2pt}{1pt}{1pt}
baseline                         & & \textbf{75.89} & \textbf{83.96} & \textbf{83.94} & 89.36 & \textbf{90.69} & \textbf{96.32} & \textbf{86.69} & 81.54 & 72.09 & 76.82 \\
\textbf{proposed} token drop     & & 75.25 & 83.64 & 83.27 & 90.00 & 90.66 & 95.20 & 86.34 & \textbf{83.33} & 71.83 & 77.58 \\
token drop + stage-2             & & 75.03 & 83.64 & 83.47 & \textbf{90.39} & 90.58 & 96.23 & 86.55 & 81.03 & \textbf{73.64} & 77.33 \\
token drop (half-rand)           & & 74.14 & 83.23 & 82.77 & 88.47 & 90.40 & 96.30 & 85.88 & 82.64 & 71.30 & 76.97 \\
token drop (half-rand) + stage-2 & & 74.61 & 83.69 & 83.20 & 89.09 & 90.35 & 95.33 & 86.04 & 82.98 & 72.42 & \textbf{77.70} \\

\bottomrule
\end{tabular}
\caption{Downstream task performance of BERT-base and BERT-large models pretrained with different input sequence lengths. Results using BERT-base and sequence length of 512 tokens are in Table \ref{tab:result-base}. Each number in the table corresponds to the average of two different pretraining and finetuning runs (using different random seeds). % \textit{Note that although the baseline method achieves the best results on many individual tasks, the averaged result shown in Table \ref{tab:summarized-results} shows that the proposed method achieves the best result overall.}
\label{tab:result-different-len}}
\end{table*}
\begin{table}[t!]
\setlength{\tabcolsep}{3.0pt}
\centering
% \fontsize{11}{13.2}\selectfont
\small
% \resizebox{\textwidth}{!}{%
\begin{tabular}{lp{3cm}}
\toprule
Methods & Average across models and downstream tasks \\
\midrule
baseline                         & \qquad \qquad 85.16 \\
\textbf{proposed} token drop     & \qquad \qquad \textbf{85.45} \\
token drop + stage-2             & \qquad \qquad 85.33 \\
token drop (half-rand)           & \qquad \qquad 85.17 \\
token drop (half-rand) + stage-2 & \qquad \qquad 85.19 \\
\bottomrule
\end{tabular}
\caption{
The averaged result of all finetuning experiments. For each method, we pretrain eight BERT models (BERT-base and BERT-large, with sequence length 128 and 512, with different random seeds), finetune them on individual GLUE and SQuAD tasks, and average all finetune results. Our proposed token dropping approach outperforms the baseline approach slightly in addition to 25\% pretraining time reduction.
\label{tab:summarized-results}}
\end{table}

\subsection{Results on Different BERT Models and Sequence Lengths}

We test our method on BERT-base and BERT-large with a sequence length of 128 and 512. We report the results in Table \ref{tab:result-different-len}. Overall, our proposed method performs similarly as the baseline method. As shown in Table \ref{tab:summarized-results}, when taking the average across all GLUE and SQuAD scores and across all four settings (two BERT models times two sequence lengths) and two pretraining runs with different random seeds, our proposed token dropping method outperforms the baseline method by 0.3\% (85.16\% to 85.45\%) in addition to the 25\% pretraining time reduction.

\begin{table}[t!]
\setlength{\tabcolsep}{3.0pt}
\centering
% \fontsize{11}{13.2}\selectfont
\small
% \resizebox{\textwidth}{!}{%
\begin{tabular}{lrr}
\toprule
Token dropping rates & GLUE-avg & SQuAD-avg \\
\midrule
drop 0\% (baseline)  & 87.19 & 84.35  \\
drop 25\%            & 87.59 & 85.23  \\
drop 50\% (\textbf{proposed}) & \textbf{87.67} & \textbf{85.77}  \\
drop 62.5\%          & 87.01 & 84.50  \\
drop 75\%            & 86.56 & 83.71  \\
\bottomrule
\end{tabular}
\caption{
Results on BERT-base models on packed sequences of 512 tokens with different token dropping rates. We see that dropping more than 62.5\% of the tokens yield worse results, whereas dropping about 50\% of the tokens yield slightly better results.
\label{tab:how_many_to_drop}}
\end{table}

\section{Related Work}
\label{sec:related}

One strategy to improve data efficiency during language model pretraining is by designing better pretraining objectives \citep{lan2020albert,clark2020electra,raffel2020t5}. Concurrently, researchers have also been exploring certain hardware properties to improve pretraining efficiency, e.g., mixed-precision training \citep{shoeybi2019megatron} and huge-batch distributed training \citep{you2020large}. 
Recently, \citet{wu2021taking} propose to tackle the efficient pretraining problem through rare words or phrases, and they provide rare words with a ``note embedding'' to make models better aware of the contextual information in a sequence. 

The faster depth-adaptive transformer approach is applied to text classification tasks \citep{liu2021faster}. It identifies important tokens by either computing the mutual information between each token and the given sequence label, or using a separate BERT model to exhaustively evaluate the masked language model loss for each token. There is a rich body of literature on faster inference of sequence generation problems, such as early layer exits during translation \citep{elbayad2020depth,han2021dynamic}, non-autoregressive machine translation \citep{gu2018nonautoregressive,tu-etal-2020-engine}, and amortizing the cost of complex decoding objectives \citep{chen-etal-2018-stable,tu-etal-2020-improving,pang2021amortized}.

Several ideas are particularly relevant to token-wise layer dropping:  \citet{zhang2020accelerating} propose to use a fixed probability to drop an entire layer during pretraining; here, we use the more fine-grained token-wise layer dropping. The dynamic halting algorithm \citep{dehghani2019universal}, motivated by the finding that transformers fail to generalize to many simple tasks, stops the processing of a token through upper layers if its representation is good enough. However, the implementation does not improve training time, as its goal is to improve performance.

\section{Conclusion}

We present a simple yet effective approach to save BERT pretraining time. Our approach identifies unimportant tokens with practically no computational overhead and cuts unnecessary computation on these unimportant tokens for training. Experiments show that BERT models pretrained in this manner save 25\% pretraining time, while generalizing similarly well on downstream tasks. We show that our token dropping approach performs better than average pooling along the sequence dimension. Future work will involve extending token dropping to pretraining transformer models that can process a much longer context, as well as extending this algorithm to a wider range of transformer-based tasks, including translation and text generation.

\section*{Acknowledgments}

The authors thank the anonymous reviewers for helpful feedback. 

\bibliography{anthology,custom}
\bibliographystyle{acl_natbib}

% \clearpage

\appendix
\section{Appendix}

\subsection{Discussion on compute}

On a high level, the FLOPs for language model pretraining come largely from the MLP layers \citep{shoeybi2019megatron}. 

Given that the attention compute grows quadratically with respect to sequence length, our approach saves $>50\%$ compute in the attention module in half of the encoder layers. But we ignore attention from our discussion, given that the FLOPs for BERT come largely from the MLP layers. In fact, the attention operations typically use smaller than $10\%$ of the total compute, and other operations like layer norm and activations are more negligible \citep{brown2020language,shoeybi2019megatron}.

The total MLP compute is proportional to $TL$ where $T$ is the number of tokens in each sequence, and $L$ is the number of total layers \citep{brown2020language,bahdanau2022flops}. In our case, given that we are dropping $50\%$ of the tokens in $50\%$ of the layers, we would save around $25\%$ of the FLOPs.

\subsection{Potential Limitations and Other Considerations}

Given that the community is paying more attention to long-document tasks \citep{beltagy2020longformer,tay2021long,pang2021quality}, it is worth investigating whether token dropping can be used to pretrain transformers with a much larger context length, like Longformer encoder decoder (LED) \citep{beltagy2020longformer} which accepts a context length of 16,384. 

One limitation is that our pretraining corpus and the downstream task datasets are in English. There is no guarantee that the same token dropping ratio applies to corpora or tasks in all other languages.

% https://medium.com/@dzmitrybahdanau/the-flops-calculus-of-language-model-training-3b19c1f025e4

\end{document}